\documentclass{article}
\usepackage{graphicx}
\title{Optimizing GoTools' Search Heuristics using Genetic Algorithms}
\author{Matthew Pratola \\ mp00aa@sandcastle.cosc.brocku.ca \\
Department of Computer Science, Brock University, Canada \\ 
and \\
Thomas Wolf \\ 
twolf@abacus.ac.brocku.ca \\
Department of Mathematics, Brock University, Canada}

\begin{document}
\maketitle
\begin{abstract}
GoTools is a program which solves life \& death problems in the game
of Go.  This paper describes experiments using a Genetic Algorithm to
optimize heuristic weights used by GoTools' tree-search. The complete
set of heuristic weights is composed of different subgroups, each of
which can be optimized with a suitable fitness function. As a useful
side product, an MPI interface for FreePascal was implemented to allow
the use of a parallelized fitness function running on a Beowulf
cluster. The aim of this exercise is to optimize the current version
of GoTools, and to make tools available in preparation of an extension
of GoTools for solving open boundary life \& death problems, which
will introduce more heuristic parameters to be fine tuned.
\end{abstract}

\section{Introduction}
The game of Go is difficult from any computer science point of view.
It allows too many moves in each position in order to be solved by
brute force search. At the same time, currently known techniques in
Artificial Intelligence (AI) are not intelligent enough to cope with
recognizing and trading qualitatively different assets quickly and
accurately while reasoning on different levels of abstraction. The
situation is further complicated by the length of the game, which
allows a good player to evaluate whether his opponent understands what
is going on, or whether the opponent's play only follows schematic
rules. 

GoTools is a specialized program that currently focuses on
solving closed-boundary life \& death problems in Go, which is more
attainable using existing techniques. Specifically, a tree-search is
employed to solve a given life \& death problem and find its status,
including statements about the type of Ko encountered, if that occurs.
However, even with a closed problem, the size of this tree can easily
become quite large and time consuming to solve.

Future plans for GoTools include extending the program's capability to
open-boundary problems.  In this future extension of the software, a
much improved heuristic will be essential as the number of potentially
useful moves becomes larger.  Such an improved heuristic will have
more parameters subject to fine tuning than the current version.  
For that purpose, genetic learning will be employed.  The
work described in this paper is meant to provide the genetic learning
tool to enable this future work, as well as to gather implementation
experience and improve the current version of GoTools.

In order to improve the current version, we look at optimizing the
heuristics used in the tree-search. A tree-search is clearly faster if
winning moves are tried first at each board position, instead of
trying a losing move first and having to learn the correct move from
a sub-tree-search.  The heuristic in GoTools that ranks different
moves before they are executed in a depth-first search is based on
parameters which we want to fine tune to improve the search speed.
The accuracy of the search is not influenced by these parameters.

Other heuristic weights which govern the pruning of the search are
able to speed up the program by sacrificing accuracy.  In this case,
we may not always solve the problem correctly, but aim to find a set
of heuristic weights which have a large improvement in speed with only
a minimal effect on solution accuracy.

Section 2 gives an overview over the different
heuristic weights, different fitness functions and the complete
environment to be used. Section \ref{RESULTS} reports the results.

\section{Overview}

GoTools has a set of rules that enable it to quickly solve a wide
variety of closed-boundary life \& death problems.  Some of these rules
are hard-coded into the program, and are always used when evaluating
a given problem.  However, some of these rules are governed by
heuristic weights - numerical parameters which can emphasize (or
de-emphasize) the effect of that rule in solving the problem.  Hence,
we refer to this subset of rules as {\em heuristic rules}.  We wish to
optimize their heuristic weights in order to speed up the solving of
problems.

In order to perform these optimizations, a large number of problems must be available
to learn from.  The problems available from the GoTools distribution are randomly generated
on a computer.  This is important as we expect the generated problems to be free of bias.  As such,
a variety of Go positions will be considered during training.  This leads to a more balanced problem solver,
making GoTools more flexible and competitive.

\subsection{Genetic Algorithms} \label{section_ga_s}
In order to optimize the heuristic weights, and hence the usefulness
of our heuristic rules, a Genetic Algorithm (GA) was implemented to
search for the best set of heuristic weights (while we do introduce
some of the relevant GA terminology for clarity, the reader unfamiliar
with the design and terminology of Genetic Algorithms is directed to
\cite{Koza}).  As we have many heuristic rules available, we also have
many heuristic weights to consider.  A set of these heuristic weights
is referred to as a {\em chromosome}.  Thus, each chromosome is a
candidate solution, containing a set of heuristic weights (whose
individual values are referred to as {\em allele's}) which we must
optimize. A Genetic Algorithm is a search technique modelled on
biological systems. It works by having a population of chromosomes,
creating new chromosomes that are in some way descendants of existing
fit chromosomes from the previous population, and removing old
chromosomes which have a low fitness.  Each occurrence of selecting
fit chromosomes, creating new child chromosomes and removing
low-fitness chromosomes constitutes one generation of the GA.  By
allowing the GA to run for many successive generations, the
chromosomes gradually adapt to the needs of the fitness function, and
better solutions are found.  Depending on how useful these heuristic
weights (encoded in a given chromosome) are in solving live \& death
problems, the chromosome is allocated a fitness value, or simply a
{\em fitness}. The function computing these values is called a {\em
fitness function}. Other factors relevant to the design of a GA is the
initial number of chromosomes known as the {\em population size}, the
number of new chromosomes, or {\em children}, created in each
generation, the total number of generations that the GA will run for,
and the function that decides which chromosomes to keep (or discard)
after each generation, the {\em selection function}.  The aim of the
search is to find an optimized chromosome with a fitness value as high
as possible in the hope that running GoTools with these heuristic
weights will solve arbitrary closed boundary life \& death problems as
fast as possible.

\subsection{Different sets of heuristic weights}
Apart from hard-coded heuristic rules (see \cite{TW00}), there are
additional rules governed by 72 parameters.  These parameters are mainly weights
that assign how trustworthy the different rules are.
 
These values can be grouped into three subsets: a static set (46 parameters), a
dynamic set (10 parameters) and a pruning set (14 parameters). As the heuristic rules corresponding to
these three sets are essentially independent of each other, their
weights are optimized in separate runs of the Genetic Algorithm, using
different chromosome sizes and partly different fitness functions.  

The purpose of our experimentation was threefold:
\begin{itemize}
\item to determine how much improvement in execution speed can be
  realized for the current version of GoTools with an optimized set of
  parameters,
\item to evaluate several training sets of varying size, i.e.\ varying the number
of problems and average difficulty level of the problems,
\item to check the feasibility of using the much faster static
  fitness function in place of the dynamic fitness function (explained below) at least for
  the optimization of static parameters.
\end{itemize} 

\subsubsection*{The set of static weights}
A subset of 46 parameters exclusively features the board position that is under investigation.  
Examples are the bonuses to be given for a move:
\begin{itemize}
\item if it falls on a potential eye-point and has a distance 2 to the
  edge of the eye (i.e.\ if the move is useful to split the eye into two
  when played by the side that wants to live or to prevent splitting
  the eye if it is played by the killer side),
\item if it completes one or more eyes,
\item if it splits 2 weak chains,
\item if it falls on a 1-2 point,
\item and others, see \cite{TW00}.
\end{itemize}

These heuristics are still relatively straight forward and only a few
are conditionally linked. As these parameters relate only to the
current board position, they are called {\em static} parameters or
weights henceforth. To train the genetic learning of static
parameters, the evaluation function can be simple and perform only the
heuristic procedure itself.  For a given set of static parameters
(i.e.\ a given chromosome), a set of life \& death problems is
evaluated.  For each problem the possible first moves are ranked by
the heuristic procedure.  Dependent on the place of the unique winning
move of the problem in this ranking, the chromosome obtains a higher
or lower fitness value.  We call this fitness function {\em static}.  It is described
in more detail further below.

A second possible fitness function {\em solves} a set of problems for 
each chromosome, and is therefore called {\em dynamic} in this paper. 
This is similar to the method used for dynamic parameter
optimization, which is described below.

\subsubsection*{The set of dynamic weights} 
The second group of parameters contains 10 heuristic weights which govern the
dynamic learning capabilities of GoTools.  Examples are:
\begin{itemize}
\item a bonus for moves that were frequently found earlier to be
winning moves,
\item a penalty for a move that in this situation is useless or
forbidden for the other side,
\item bonuses for what were favored moves for the other side in the
previous position.
\end{itemize}

Because parameters in this group weight the relevance of information
gathered during the tree-search performed so far, we will call them
{\em dynamic} parameters or weights. The evaluation function used in
the genetic learning of dynamic parameters cannot simply perform a
static heuristic for the original position of the life \& death
problem, these problems have to be solved. Depending on the difficulty
of the life \& death problems, it can be 100 - 100,000 times more
expensive to solve them than to perform an initial heuristic.
In order to meet the increased computation time requirements, a Beowulf 
\cite{Beowulf}
cluster is used where different chromosomes are evaluated on different
slave nodes in parallel, each chromosome being used to solve a training set of problems.
Because GoTools and its infrastructure are written in Pascal using the
FreePascal ({\tt http://www.freepascal.org}) compiler,
a Pascal interface to the Message Passing Interface (MPI) library was written.  
It is available freely under {\tt http://lie.math.brocku.ca/twolf/htdocs/MPI}.
The Genetic Algorithm was adapted accordingly to make use of it.

Initial testing regarded the optimization of the GA itself.
We chose 22 children per generation as our Beowulf cluster has 24 nodes - one
is used as a master node and we wanted an even number of children.
For the crossover rate, we chose a value of $C_{0}=6.5\%$, while the probability for a chromosome
to undergo mutation was $M_{0}=50\%$.

To measure the progress in optimizing the dynamic parameters, we
compare the optimized set of values with the ones currently used in
GoTools and also with a set of values identically zero where any dynamic
rules are effectively disabled. Also, for this evaluation, a
problem set called the test set is used, which is different from the training set.

\subsubsection*{Pruning parameters}

Although we are not going to optimize the following set of parameters in this paper,
we will still describe it for completeness.
The version of GoTools as it is currently (2002) operating under 
\linebreak {\tt http://lie.math.brocku.ca/GoTools/applet.html} allows
5 speed levels: from the exact and slowest mode 1 to the fastest mode 5,
each mode providing a speed up of a factor of roughly 2.
The pruning of the
search tree is decided by rules (see \cite{TW00}), where each of them
can be used more or less aggressively, depending on the so-called 
{\em pruning} parameters. 

\begin{figure} \begin{center}
\includegraphics[width=10cm,height=8cm]{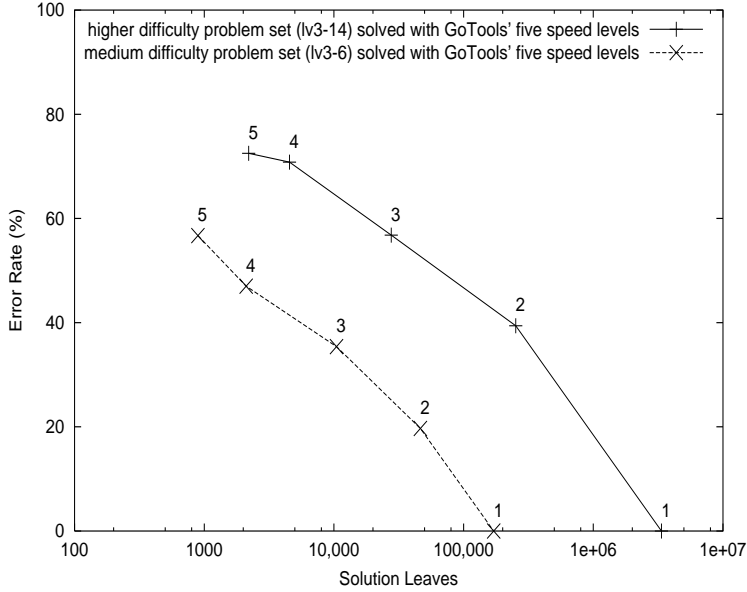}
\caption{Error rates of the current version of GoTools as a function
of its speed as characterized by the total number of leaves of all
problems in the high difficulty test set (lv3-14).}
\label{prune-default}
\end{center}   \end{figure}

It is desirable to have different sets of pruning parameters
which cover a wide range of speed up levels and each set of pruning
parameters being error-minimized for its level of speed up.  

In Figure \ref{prune-default}, the dependence of the error rate on the
number of terminal leaves is shown for the 5 accuracy levels and
two difficulty levels.  
Error rates are relatively high, as a problem was already considered to be
wrongly solved 
if the type of ko was not correctly determined. 

We suggest to use these curves to construct a fitness function.  The quality
of a set of pruning parameters could be judged by comparing its error rate with the
error rate in Figure 1 given its speed, i.e.\ number of leaves.
The genetic learning of optimal pruning parameters will be the object 
of future work.

\subsection{Measuring Performance}
In optimizing the execution speed of GoTools, there are primarily 2
ways we can measure our progress: by counting the tree-search's
terminal leaves, or by measuring the wall clock time required to solve
problems.  Our preference is to measure terminal leaves for many
reasons.  First, this measurement is consistent across different CPU's
or even across different nodes in our cluster of identical machines.
Second, this is a more natural measurement for which the evaluation
function of our Genetic Algorithm is based on.  Finally, it is also
more relevant to see a reduction in a problem's search tree size as a
measure of heuristic functionality than a more abstract measurement
such as wall clock time.  However, is it valid to claim that smaller
solution trees also execute faster?  In the current version of
GoTools, this is indeed the case as all the heuristic rules are of
similar algorithmic order.  However, we can make a simple test to
convince us that this is the case.

We ran a high difficulty level test set with approximately 250
problems using 2 heuristics (the current GoTools heuristic, and our
newly optimized heuristic - see Section \ref{RESULTS}).  We collected
the execution times and solution leave counts over $n_o=n_n=n=10$ runs for
each heuristic. That is, we have $L_{orig}$ leaves from our current heuristic,
$L_{new}$ leaves from our new heuristic and 2 arrays of execution
times, $(T_o^1,T_o^2,...,T_o^{10})$ and 
$(T_n^1,T_n^2,...,T_n^{10})$ for our original and new
heuristics respectively. With the following short calculation we want
to show that the mean execution time per leave is equal.  For that, we compare the time per leave
$\tau_o^i=T_o^i / L_o$ for the original heuristic with the time per
leave $\tau_n^i=T_n^i / L_n$ for the new heuristic and will find that
the averages are equal. We denote the averages with a bar, 
e.g.\ $\overline{\tau_o}=\frac{1}{n}\sum_{i=1}^n (\frac{T_o^i}{L_o})$.
Analysis of our collected normalized execution times $\tau_o, \tau_n$ 
indicates that these values are normal independently distributed 
with equal variances, so we can
perform a difference of means test as follows \cite{statstext}:
\[(\overline{\tau_o} - \overline{\tau_n}) - t_{\mbox{\scriptsize student}} 
< \mu_{\tau_o} - \mu_{\tau_n} < 
  (\overline{\tau_o} - \overline{\tau_n}) + t_{\mbox{\scriptsize student}} \]
where 
\begin{eqnarray*}
t_{\mbox{\scriptsize student}} & = & t_{\frac{\alpha}{2}, n_o+n_n-2} S_p \sqrt{\frac{1}{n_o}+\frac{1}{n_n}}\\
 \mu_{\tau_o} - \mu_{\tau_n} & = & \mbox{difference of the true averages of $\tau_o, \tau_n$} \\
S_p & = & \sqrt{\frac{(n_o-1)s_o^2 + (n_n-1)s_n^2}{n_o+n_n-2}} \\
s_o^2 & = & \frac{\sum_i^n (\tau_o^i - \overline{\tau_o})^2}{n_o-1} \ \ \ (\mbox{variance of $\tau_o$}) \\
s_n^2 & = & \frac{\sum_i^n (\tau_n^i - \overline{\tau_n})^2}{n_n-1} \ \ \ (\mbox{variance of $\tau_n$}) \\
t_{\frac{\alpha}{2}, n_o+n_n-2} & = & t(n_o+n_n-2) \ \ \ (\mbox{Student t distribution with}    \\
                                &   & \ \ \ \ \ \ \ \ \ \ \ \ \ \ \ \ \ \ \ 
                                      \ \ \ \ \mbox{$n_o+n_n-2$ degrees of freedom}) 
\end{eqnarray*}
Analysis conducted at the 95\% confidence level ($\alpha=0.05$)
resulted in the following confidence interval for the difference of
means: \[(-8.5428*10^{-7},1.747*10^{-5}) .\] Since the point $0$ is included within our
confidence interval, we can conclude that there is no significant
difference between execution times per terminal leave for different
heuristics at the $\alpha=0.05$ level of significance, and that the counting of terminal leaves is therefore a
valid performance measurement.

\subsection{Experimental Procedure}

Experimentation with the Genetic Algorithm was done using the following general framework:
\begin{enumerate}
\item Run the Genetic Algorithm
      on a training set of problems.
\item Save the optimal chromosome that has been found, i.e.\ 
      the set of optimal heuristic weights.
\item Run
      a test program with optimized heuristic weights by solving a
      test set of problems which are different from the training set.
      Use test sets of easy, medium and high difficult problems.
\item Record solution leaves reported
      in these runs.
\item Repeat the whole procedure with a training set of a different
      size and later with a training set of different difficulty level.
\end{enumerate} 

Both heuristic sets (static and dynamic) are optimized independently using for the other currently not optimized set the original values.
The {\em heuristic-test} program allows the tester to quickly run any new
heuristic weights on a problem set to find the resulting speed-up in execution.
Training and testing can be done on a large set of data, as the current GoTools
library consists of 6 volumes of problems, each volume being
sub-divided into 14 levels of difficulty with roughly 280 problems each.  These problems are stored
in files named lv{\em A}-{\em B}, where '{\em A}' represents the
volume number from 1 through 6, and '{\em B}' represents the
difficulty level, enumerated from 1 through 14 in increasing
difficulty.  Problems for the training set were taken from
the files lv3-6, lv3-10 and lv3-14, and test set problems came from lv4-6, lv4-10 and lv4-14, all contained in the GoTools distribution.

\subsection{Genetic Algorithm Implementation}

Genetic Algorithms are well adapted to performing large
searches and attaining near-optimum results when designed well in
accordance with the problem. The
heuristic weights governing the tree-search are the primary values we
wish to optimize with the Genetic Algorithm.  The GA utilized was
implemented to support two different fitness functions, one static 
and one dynamic, as explained in Section \ref{section_ga_s}.  In order to speed-up the dynamic
fitness function which solves the problems given the training set, it was executed
in parallel on a Beowulf cluster using the Message Passing Interface (MPI).

\subsubsection{``Static'' Fitness Function} \label{Static_Heuristic_GA}
The static fitness function executes very fast as it only computes a heuristic ranking
of the possible moves in the original problem position without performing a tree-search.
The fitness value depends on where in this ranking the unique winning move appears.  The
unique winning move is read together with the problem from a training set of problems.
The pseudo code for the static GA is:

\noindent
\vspace{6pt}\\
\hspace*{12pt} Initialize GA \\
\hspace*{12pt} {\bf while} \\
\hspace*{12pt} \hspace*{12pt} Select parents for reproduction \\
\hspace*{12pt} \hspace*{12pt} Create children through crossover and mutation operators \\
\hspace*{12pt} \hspace*{12pt} Apply fitness function to evaluate children \\
\hspace*{12pt} \hspace*{12pt} Select parents to be replaced by new generation \\
\hspace*{12pt} {\bf until} stopping condition \vspace{9pt}\\

Tuning the GA for good performance typically involves selecting a good
crossover rate, mutation rate and above all, a good fitness function.
Initial testing indicated that good GA performance was best realized
with a reasonably high mutation rate and a fairly low crossover rate.
This is indicative of the integer-valued heuristics that
have a wide range of possible values, but also a fairly low degree of
correlation between individual heuristics. Each weight was allowed to
vary in the interval [0,1000].

\begin{figure} \begin{center}
\includegraphics[width=10cm,height=8cm]{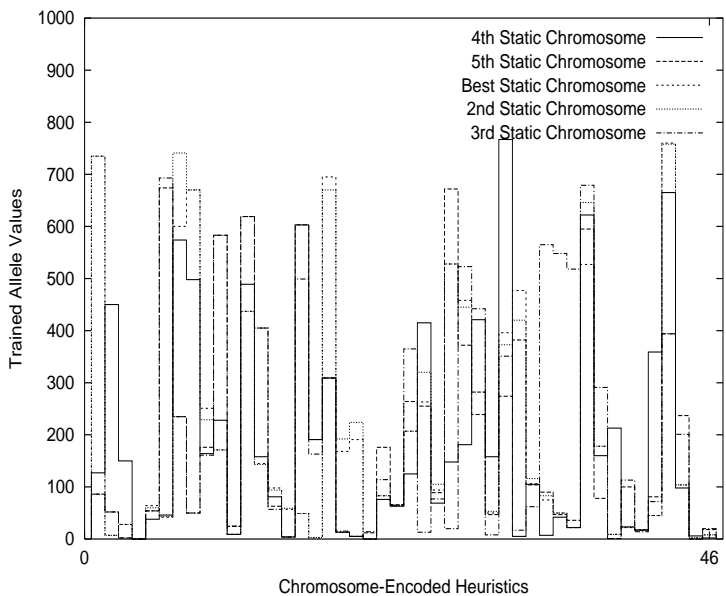}
\caption{Top 5 static chromosomes with allele values trained using a static fitness function.} 
\label{static_heuristics}
\end{center}   \end{figure}

Figure \ref{static_heuristics} shows that some optimal allele values
have to be low, others have to be high and again others vary
noticeable. We selected a crossover rate of $C_{0}=6.5\%$ and a
mutation rate of $M_{0}=50\%$ for all subsequent testing.  As such,
for each generation of the GA, there is a 6.5\% chance of the
crossover operator being applied to a chromosome, and a 50\% chance of
the mutation operator being applied to the chromosome.

Varying the fitness function of the static GA is fairly simple as it
only involves changing the five bonus weights present, i.e.\ the bonus
given if the unique best move comes $1^{\mbox{\scriptsize st}},
2^{\mbox{\scriptsize nd}}, 3^{\mbox{\scriptsize rd}},
4^{\mbox{\scriptsize th}}$ or $5^{\mbox{\scriptsize th}}$ in the
ranking of moves of the heuristic.  A few different tuples were
evaluated, with linearly, polynomially and exponentially
falling values;
in the end, a rather arbitrary
tuple \begin{math}\langle 20,13,7,3,1\rangle
\end{math} which exhibited good performance was selected.

The selection of population size and number of children proved to be
very important with the static GA.  For the easiest problems, lv3-6, a
population size of 100 with 80 children per generation achieved good
results, higher values having diminishing returns.  However, as the
difficulty increased, the resulting heuristics worsen considerably
unless the population size and number of children per generation are
incremented accordingly.  This reduces the training-time speed-up that
the static GA was hoped to allow.

\subsubsection{``Dynamic'' Fitness Function} \label{dyn_heuristic_implementation}
Optimizing the dynamic heuristic values is difficult as the execution
time to run the GA is quite long, and increases rapidly as the
difficulty level of the problems is increased.  Consider optimizing
the dynamic heuristics with a training problem set containing 200
problems, a GA population and children size of 20 chromosomes with only 30
GA generations.  This means that 120,000 problems must be fully solved
by GoTools in this case.  As the difficulty level increases, the time
to perform this search quickly exceeds the capability of a single
workstation to solve these problems in a reasonable time frame. To
help reduce this effect, the GA was parallelized using the Message
Passing Interface (MPI) (see \cite{MPICH} and \cite{MPIintro}).

A serial version of the GA would typically take many hours to execute
for simple problems, to many days for difficult problems.  Conversely,
the parallel version is able to run in the range of tens of minutes to
many hours for the same difficulty range.  For instance, if we now run
the above example across 20 CPU's, each CPU is now only responsible
for solving 6,000 problems which equates a 95\% reduction in GA
execution time, in the best case.  This speed-up was crucial in
enabling the tuning of GA parameters and in searching and testing
different heuristic sets within a reasonable amount of time.  It also
serves as a first-run implementation of a pipelining architecture that
can handle large population sizes on a given, small number of
available computation nodes, which may prove important in future
work on GoTools.

The pseudo code for the dynamic GA is shown below. \vspace{12pt}\\
\noindent
\hspace*{12pt} Master Node: \vspace{6pt}\\
\hspace*{12pt} Initialize GA \\
\hspace*{12pt} {\bf while} \\
\hspace*{12pt} \hspace*{12pt} Select parents for reproduction \\
\hspace*{12pt} \hspace*{12pt} Create children through crossover and mutation operators \\
\hspace*{12pt} \hspace*{12pt} Send children to slave nodes \\
\hspace*{12pt} \hspace*{12pt} Evaluate children (in parallel on slave nodes) \\
\hspace*{12pt} \hspace*{12pt} Receive evaluations from slaves \\
\hspace*{12pt} \hspace*{12pt} Select parents to be replaced by new generation \\
\hspace*{12pt} {\bf until} stopping condition \vspace{9pt}\\
\hspace*{12pt} Slave Nodes: \vspace{6pt}\\
\hspace*{12pt} {\bf while} \\
\hspace*{12pt} \hspace*{12pt} Receive child from master node\\
\hspace*{12pt} \hspace*{12pt} Apply fitness function to evaluate child \\
\hspace*{12pt} \hspace*{12pt} Send Evaluation to Master \\
\hspace*{12pt} {\bf until} stopping condition \vspace{12pt}

\begin{figure} \begin{center}
\includegraphics[width=10cm,height=8cm]{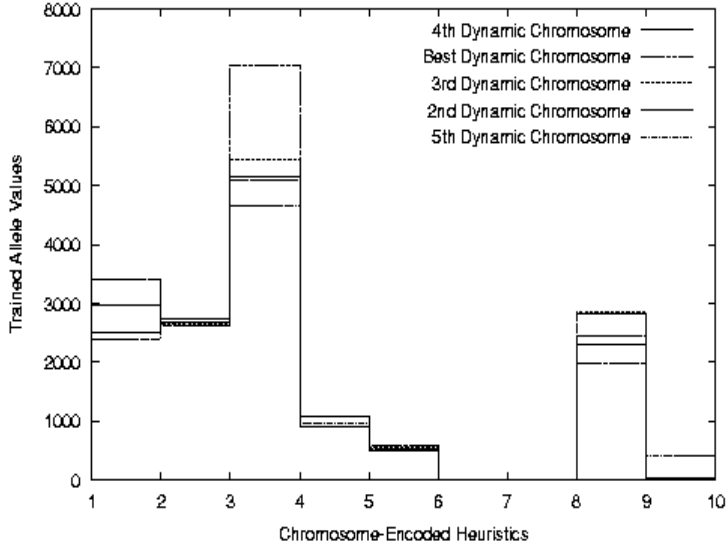}
\caption{Top 5 dynamic chromosomes with allele values trained using dynamic fitness function.} 
\label{dyn-heuristics}
\end{center}   \end{figure}

Tuning the GA for good performance typically involves selecting a good
crossover rate, mutation rate and above all, a good fitness function.
The former proved relatively easy, while the latter proved to be more
difficult.  Similar to the static GA, a reasonably high mutation rate
and a fairly low crossover rate were selected.  We selected a
crossover rate of $C_{0}=6.5\%$ and a mutation rate of $M_{0}=50\%$.
Heuristic weights were randomly initialized in the range of [0,10000]
and the resulting top 5 chromosomes are shown in Figure \ref{dyn-heuristics}.

A striking feature is the high variation of a single heuristic weight across different
top chromosomes.  The possible reasons for this observation include:

\begin{itemize}
\item strong correlation with a large variation in another optimal allele value,
\item our training set is too small,
\item a heuristic rule is used only rarely, and if it is applied, then
it should dominate other rules, although by how much is not relevant.
\end{itemize} 
Which of these explanations is appropriate would have to be studied for each
of the alleles individually, but at least this diagram gives us some useful
information as to which heuristic rules show high variation weights.
 
We also notice that some optimal heuristic weights have a large value
while others have very small values (such as heuristic 6).  This
indicates that some heuristic rules appear to provide little
additional information.  Whether this is truly the case is an area for future study.

For each problem solved in the training set, GoTools returns the number of terminal leaves
from that problem's tree-search.  How should these numbers be combined to calculate a fitness for all the problems in the training set?
There are two ways in which we may construct a total fitness value:

\begin{enumerate}
\item all problems are given equal weight.
\item hard problems are favored over easy problems.
\end{enumerate} 

GoTools is already fast at solving easy problems, so it is natural to
emphasize our optimization on the harder problems so we may solve them
quicker in practice.  Therefore, the performance of hard problems
should have a greater influence on the fitness value.  The
measurements available, from which we must construct this fitness
function, are the number of terminal leaves ($tl_{new}$) of the current chromosome,
and the number of terminal leaves ($tl_{old}$) of a reference chromosome containing the pre-optimization heuristic weights.
Several fitness functions were investigated, the four main ones are shown below:

\begin{list}{}{}
\item \begin{equation} \mbox{fitness}= 
      \sum_{\mbox{\scriptsize\ training problems}}\frac{(\mbox{tl}_{\mbox{\scriptsize\ old}})^{2}}
       {\mbox{tl}_{\mbox{\scriptsize\ new}}} 
      \end{equation}
\item \begin{equation} \mbox{fitness}= 
      \sum_{\mbox{\scriptsize\ training problems}}\frac{\mbox{tl}_{\mbox{\scriptsize\ old}}}{\mbox{\mbox{tl}}_{\mbox{\scriptsize\ new}}}
	\end{equation}
\item \begin{equation} \mbox{fitness}= 
  \sum_{\mbox{\scriptsize\ training problems}} -\mbox{tl}_{\mbox{\scriptsize\ new}} \end{equation}
\item \begin{equation} \mbox{fitness}= 
      \sum_{\mbox{\scriptsize\ training problems}}\frac{1}{\mbox{\mbox{tl}}_{\mbox{\scriptsize\ new}}}
	\end{equation}
\end{list}

After calculating the raw fitness values for each chromosome based on
all problems in the training set, the fitness values of all chromosomes are shifted and
linearly scaled to fit in the interval [0,1] so that the values are
normalized for the GA's selection function.  With fitness functions
(1), (2) and (3), the performance in solving harder problems has a
higher impact in the fitness value. This
behavior is desired as GoTools is already well optimized for small
problems.  Function (4) was included in our evaluation for reference
purposes, as it does not favor hard problems.  Comparisons between
the four fitness functions showed that function (3) provides a good
balance between larger and smaller problems.  All results reported
below were obtained using this fitness function.  In principal, there is a danger
of overemphasizing hard problems because their search tree is exponentially larger than the one
of a simple problem.  By counting the leaves of the search tree (i.e.\ measuring its surface)
instead of counting all nodes (i.e.\ the volume of the search tree), we reduce this
tendency of much larger values for slightly harder problems.

Finally, one must select the population size and number of children of
each generation.  Increasing the population size did not provide
additional improvement as the heuristics are weakly linked and there
are only ten dynamic heuristics to consider.  Further, it was found
that the best results were obtained with the number of children
ideally matching the population size.  Tests showed that population
sizes and number of children on the order of 20 were most appropriate;
since 24 nodes are available on our Beowulf cluster, we settled for 22
chromosomes forming the population size and number of children per
generation (our parallel GA always requires an even number of children
, plus the root node, thereby
fully using our 24 node cluster).  Our GA program is easily adjustable
to support various population sizes, number of problems in the
training set and the number of processors available to perform
computations.

\section{Results} \label{RESULTS}

Optimization of both the static heuristic weights and the dynamic
heuristic weights provided improvements in search performance.

The static heuristic weights trained with the static fitness function
did not provide any tangible improvement, and the results obtained
indicate that a GA using this fitness function is only good at
improving fitness weights, but not in solving other test sets of
problems. However, static heuristic weights trained with the dynamic
fitness function performed much better, giving an improvement of
around 12\% from the baseline (un-trained weights).

With a trained dynamic heuristic, an improvement of around 18\% from
the baseline (untrained weights) was achieved over many different
types of problems.  The optimized dynamic heuristic provides an 8\%
reduced terminal leaf count compared with that of the original
heuristic weights.  When combined with our trained static heuristic
weights, the overall improvement is around 20\%.  
These results are discussed in more detail in the following subsection.

The relatively moderate improvement achieved by optimizing heuristic weights
is interpreted as follows. The strength of GoTools comes mainly from
an early life and death detection, the use of a hash database to learn
intermediate search results and other hard-wired learning mechanisms.
The collection of heuristic rules is comparatively underdeveloped.
Hence an improvement of weights can only have a limited value.  The
work to be done on improving the heuristic rules themselves will be
much simplified with the genetic learning tools at hand which not only
can fine-tune parameters but can also be used to judge the quality of the
heuristic rules through a comparison of the achieved efficiency.

\subsection{Static GA} \label{StaticGA}

Optimizing the static heuristics using only a static fitness function
proved to be difficult, as the complexity of solving life \& death
problems is largely hidden from this function. Furthermore, as problem
difficulty increases, the performance of the static heuristic
deteriorates. The reason seems to be that difficult problems with many
possible moves are not easily solved using simple heuristic rules and are
therefore also not useful as training sets to learn the weights of simple heuristic
rules. Therefore, to realize at least average results with the
static heuristic, the training time becomes increasingly large as the
population size must be increased accordingly.  This indicates that
the static heuristic function is not adequate to learn the solving of
arbitrary test sets well.

\begin{figure} \begin{center}
\includegraphics[width=10cm,height=8cm]{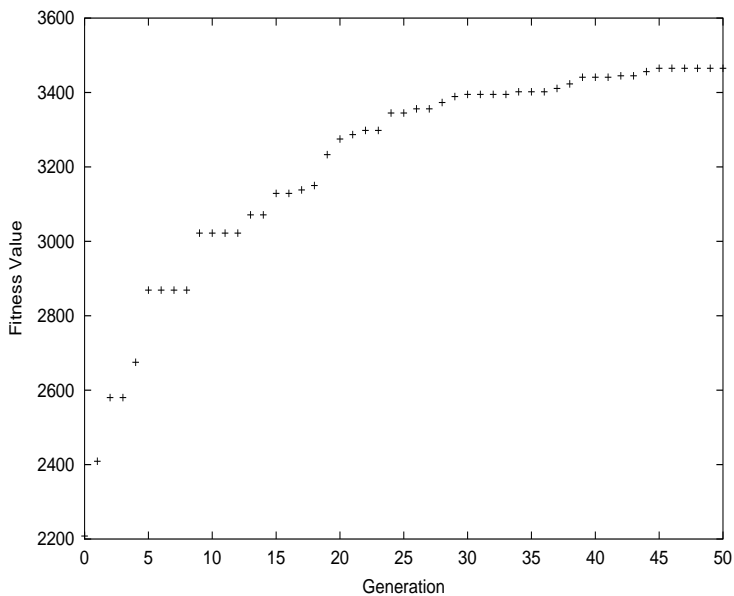}
\caption{Typical learning curve when training static weights with a static fitness function} 
\label{static_ga_learning_curve}
\end{center}   \end{figure}

\begin{figure} \begin{center}
\includegraphics[width=10cm,height=8cm]{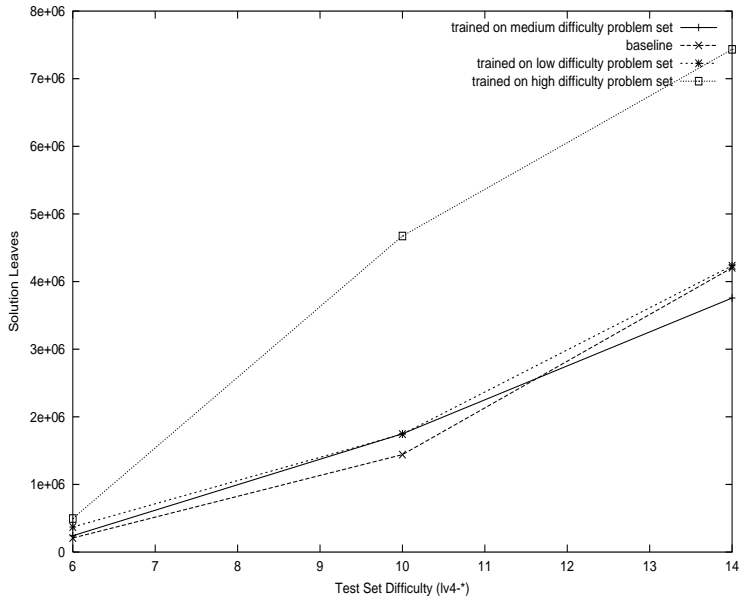}
\caption{Solution leaves of static weights when trained with a static fitness function using a training set of 128 problems from lv3-6, lv3-10 and lv3-14 with varying test set (lv4-6, lv4-10, lv4-14) difficulty level.}  
\label{static_trends}
\end{center}   \end{figure}

Figure \ref{static_ga_learning_curve} shows a typical learning curve
for the static heuristic function.  The GA was run with a population
size of 100 and a children size of 80, along with other parameters
tuned as described in Section \ref{Static_Heuristic_GA}. Table
\ref{static_training_sets_on_lv3-6} shows three runs using the static
fitness function to optimize static weights with problems from the
easy training set (lv3-6) tested on an easy test set (lv4-6) with
three different training set sizes.  The outcome that a training set
size of 128 is slightly better than 200 is only a coincidence, but
even if a training set size of 200 would have been slightly better, it
would not have justified using a training set nearly twice as large.
Based on this outcome, we used a training set size of 128 problems for
the computation show in Figure \ref{static_trends}.

In this Figure, the performance of three static heuristic sets, each
trained with training sets of varying difficulty (lv3-6, lv3-10,
lv3-14), is shown when tested with test sets of varying difficulty
(lv4-6, lv4-10, lv4-14).  The dominating feature of this diagram is
the bad quality of weights trained with the difficult training set.
Even if a larger GA population, more generations and a larger training
set size might improve the performance, it is unlikely to reach the
quality obtained with the dynamic fitness function.

\begin{table}[h]
\begin{center}
\begin{tabular}{|c|c|c|c|} \hline
{\em Number of problems in training set:} & 64 & 128 & 200\\
\hline {\em Solution Leaves:} & 370,115 & 233,037 & 234,131\\
\hline
\end{tabular} 
\caption{Solution leaves of static weights trained with a static fitness function using the training set lv3-6 and the test set lv4-6 with varying training set sizes.} 
\label{static_training_sets_on_lv3-6}
\end{center}
\end{table}

\begin{figure} \begin{center}
\includegraphics[width=10cm,height=8cm]{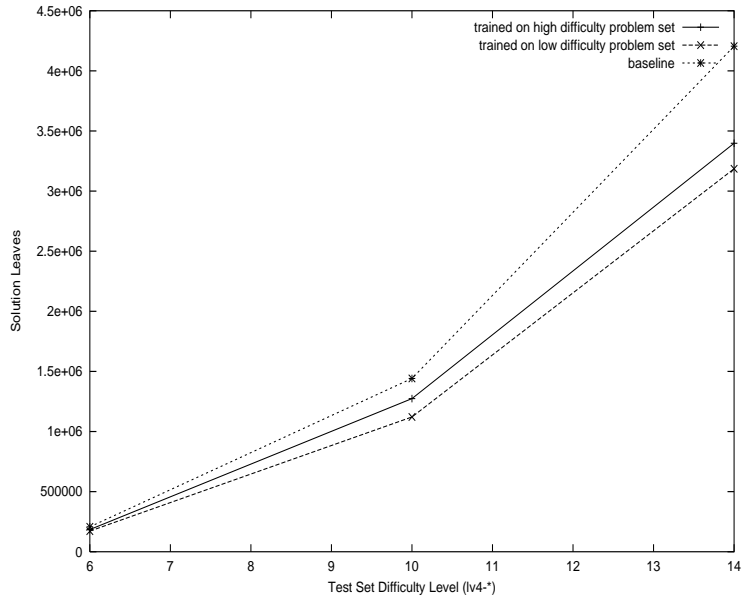}
\caption{Solution leaves of static weights when trained with a dynamic fitness function using a training set of 128 problems from lv3-6 and lv3-14 with varying test set (lv4-6, lv4-10, lv4-14) difficulty level.}
\label{static_solve_trends}
\end{center} \end{figure}

As we suspected that the static fitness function was likely limiting the optimization of the static weights, 
the same static parameters were also optimized by
completely solving the problems, that is, using the same fitness function as utilized in optimizing
the dynamic parameters (see \ref{dyn_heuristic_implementation}).  We utilized a training set of 128 problems at
various difficulty levels with a population size of 22 chromosomes and 22 children per generation.
The GA was run for only 15 generations, and all other GA parameters remain the same as discussed earlier.  This approach confirmed our suspicions, as shown in Figure \ref{static_solve_trends}.

The results shown in \ref{static_solve_trends} indicate that, with a
proper fitness function, the optimized static weights now perform 
clearly better.
Furthermore, there is another useful trend evident in this
diagram.  We can see that the performance obtained when trained with
easy problems (lv3-6) or hard problems (lv3-14) is very similar
much in contrast to training with the static fitness function.
Results obtained with a medium difficulty training set
(lv3-10) also followed this trend, but are left out for graphical
clarity.  In fact, we can see that the heuristic weights trained with
the easy problem set actually out-perform those trained with a hard
problem set by a small margin.

\begin{figure} \begin{center}
\includegraphics[width=10cm,height=8cm]{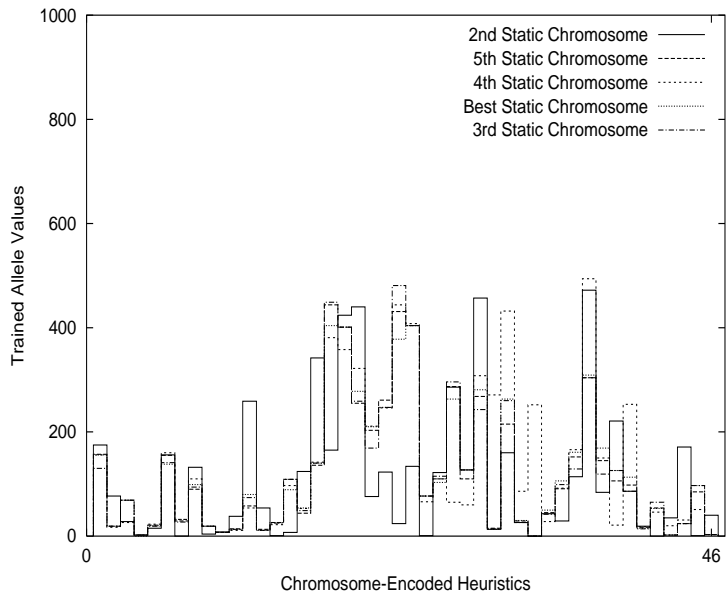}
\caption{Top 5 static chromosomes with allele values trained with dynamic fitness function.}
\label{static_solve_heuristics}
\end{center} \end{figure}

Looking at the trained heuristic weights as shown in Figure
\ref{static_solve_heuristics}, we can clearly see the difference to
Figure \ref{static_heuristics} where the variance of allele values
over the top five chromosomes is clearly higher.  Although training
with a dynamic fitness function which solves problems is much slower
than using a static fitness function, this disadvantage is partially
compensated by the fact that solving easy training sets, like lv3-6,
is much faster than using a difficult training set.  While not as fast
as using a static fitness function, solving easy training set problems
is still relatively quick on our Beowulf cluster, and much faster than
training with a high difficulty training set as is required for the
optimization of dynamic parameters, which is discussed below.

\subsection{Dynamic GA} \label{DynamicGA}

\begin{figure} \begin{center}
\includegraphics[width=10cm,height=8cm]{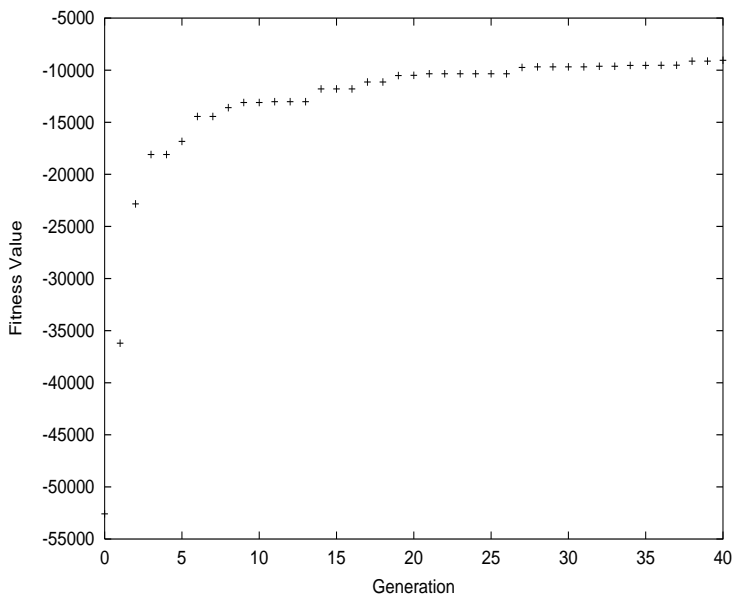}
\caption{Typical learning curve when training dynamic weights with a dynamic fitness function.} 
\label{dynamic_ga_learning_curve}
\end{center}   \end{figure}

Obtaining a good dynamic heuristic can potentially bring a noticeable
improvement to the already well-optimized GoTools tree-search.
However, optimizing the dynamic heuristics involves solving the
problem, which can quickly make such work very time consuming.  A
typical learning curve for the Dynamic GA is shown in Figure
\ref{dynamic_ga_learning_curve}.

We can see that for the training set used in Figure
\ref{dynamic_ga_learning_curve} that training with a dynamic fitness function is able to quickly
converge.  A training set of higher difficulty does tend to converge
more slowly, but in general, 30 to 40 generations were
sufficient.  Figure \ref{dynamic_training_sets_on_lv3-6} shows how the
performance of the dynamic GA varies with training set size, as shown
by the results obtained in training with 64, 128 and 200 problems taken from the full lv3-6 set.

\begin{table}[h]
\begin{center}
\begin{tabular}{|c|c|c|c|} \hline
{\em Number of problems in training set:} & 64 & 128 & 200\\
\hline {\em Solution Leaves:} & 186,005 & 171,495 & 170,633\\
\hline
\end{tabular} 
\caption{Solution leaves vs. training set size when optimizing dynamic weights using the dynamic fitness function with lv3-6.} 
\label{dynamic_training_sets_on_lv3-6}
\end{center}
\end{table}

\begin{figure} \begin{center}
\includegraphics[width=10cm,height=8cm]{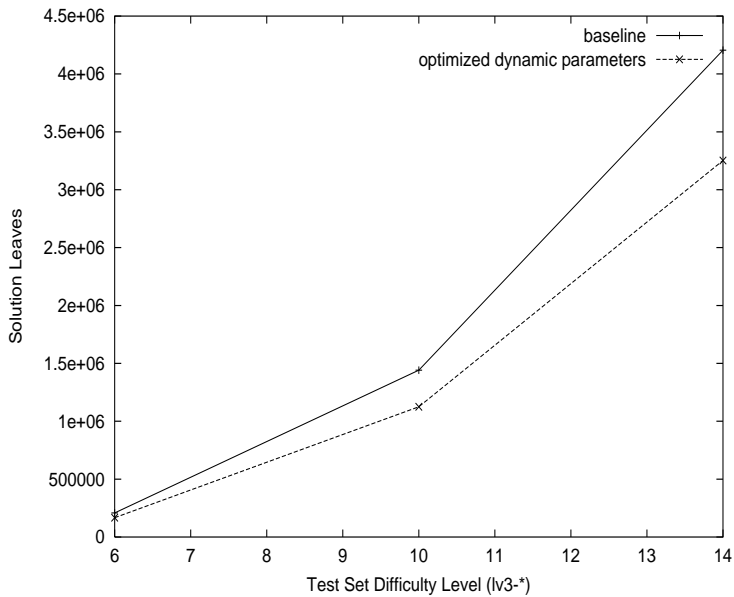}
\caption{Solution leaves for zero dynamic parameters (baseline curve) and optimized dynamic parameters using lv3-6, lv3-10
and lv3-14 as test sets.}  
\label{dynamic_vs_baseline}
\end{center}   \end{figure}

The table shows the solution leaves as reported when running a test set.
There is a reasonable improvement of
8\% achieved by increasing the size from 64 to 128 problems, while
only a minor further improvement of 0.5\% when increasing the size
from 128 to 200 problems.  When moving to more difficult problems, the
improvement between a training set size of 128 and 200 problems does
increase, however, overall a training set size of 128 problems is
favorable in terms of results and the required computation time for
training.  Subsequently, we use a training set size of 128 problems
for our main result, shown in Figure \ref{dynamic_vs_baseline}.

When optimizing static weights, it was advantageous to use easy
problems for the dynamic fitness function.  Conversely, here a
high-difficulty training set is sufficient to perform well with an
easy test set and necessary to perform well with a difficult test
set. Therefore, we have trained with the lv3-14 training set with 128
problems, as shown in Figure \ref{dynamic_vs_baseline}.  The baseline
values are obtained by using un-trained dynamic parameters.  The
results shown indicate a 14\%, 18\% and 23\% improvement in solution
leaves at testing set difficulty levels of 6, 10 and 14 respectively.
These results are consistent across many problem difficulty levels.
The only requirement is to train with high difficulty training sets.
Hence, the dynamic fitness function is preferred over the static
fitness function as while training does take longer, it is only
required once and can be well applied to a variety of problems with
good results.

\subsection{Profiling} \label{Profiling}
\begin{figure}\begin{center}
\includegraphics[width=15cm,height=8cm]{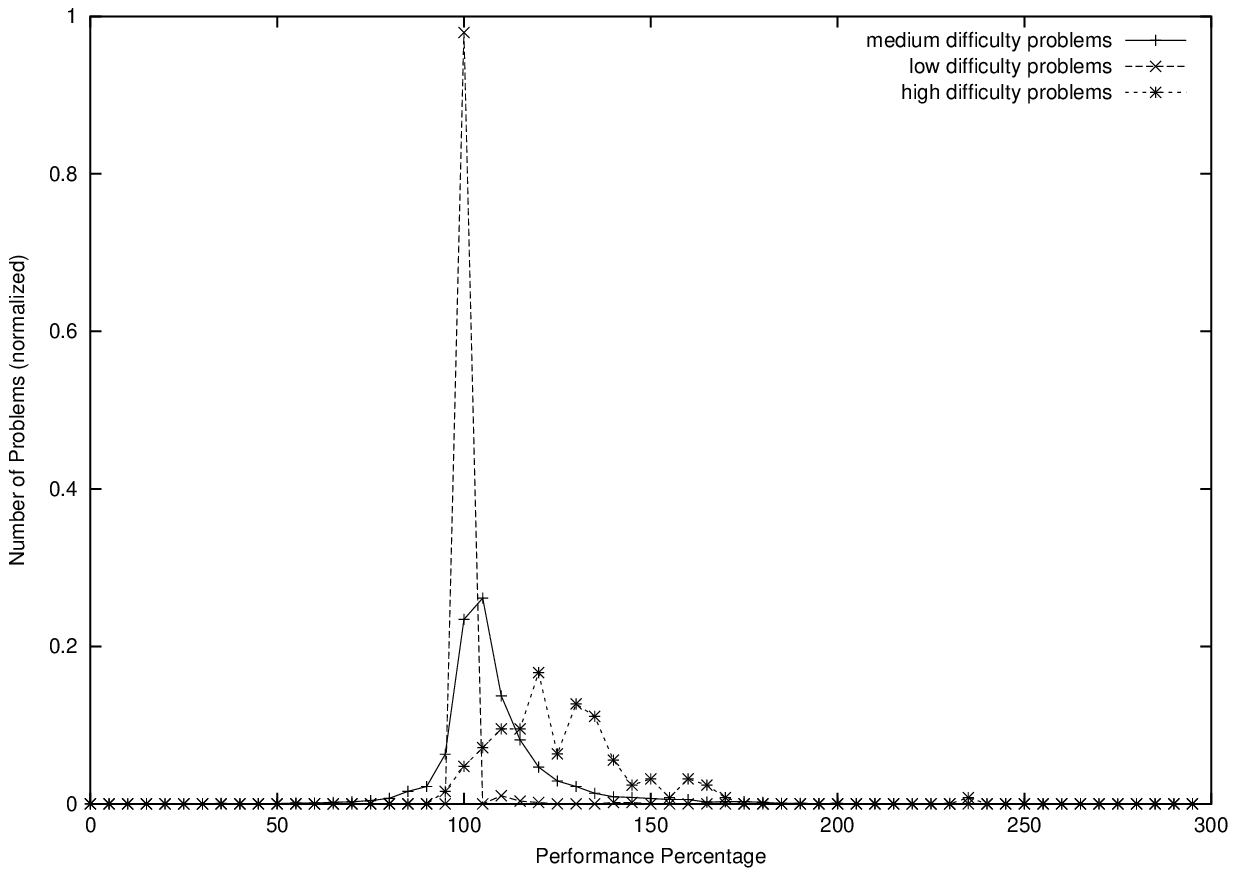}
\includegraphics[width=15cm,height=8cm]{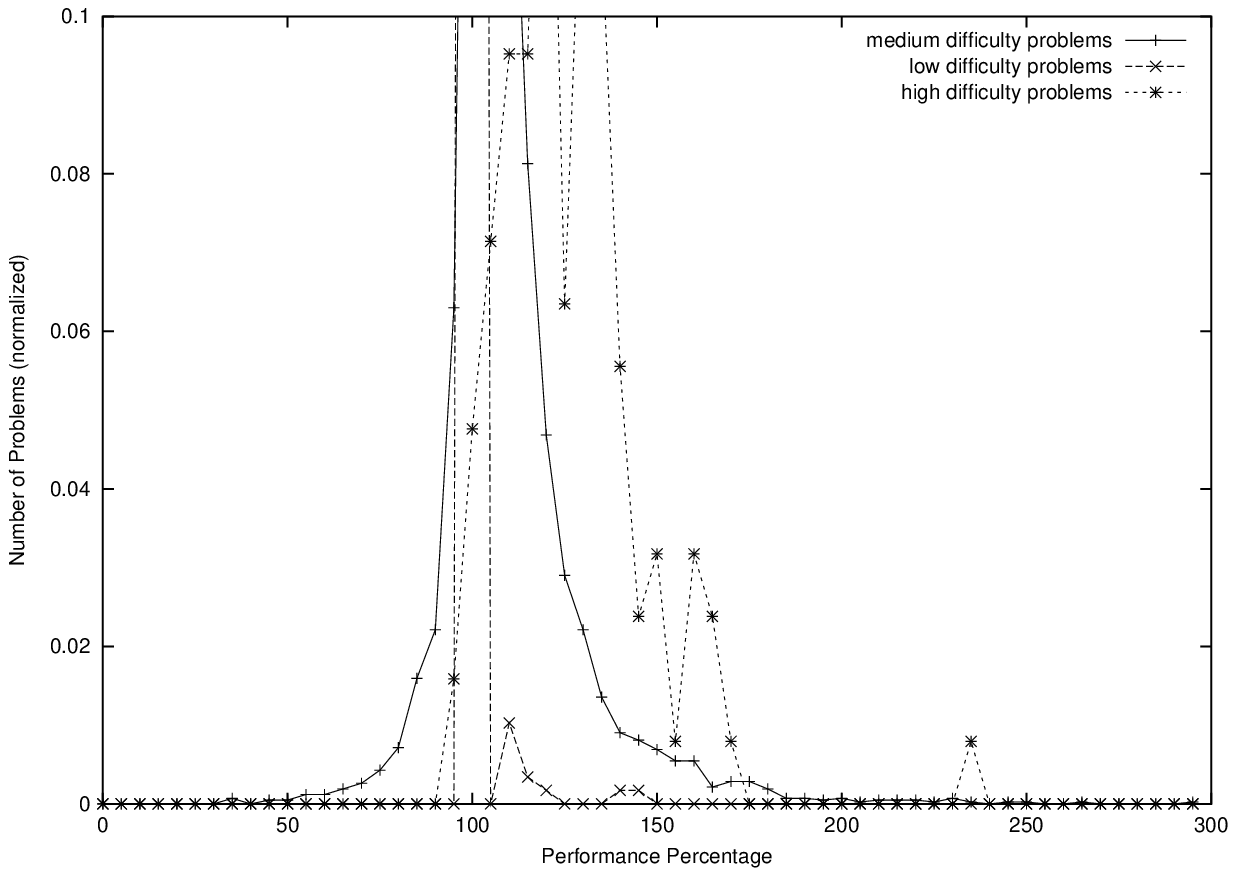}
\caption{Profile plot with baseline heuristic relative to original heuristic for easy, medium and hard difficulty level problems.}
\label{heu-none-profile}
\end{center}\end{figure}

Although the efficiency of the original heuristic weights were improved
through the genetic learning of new weights, the improvement was not
overwhelming. We now wish to analyze in greater detail the limiting
behavior of our current heuristic rules. To this aim, we generated
Figures \ref{heu-none-profile} thru
\ref{heu-iamspecial-profile}, which we refer to as {\em profile
plots}.

\begin{figure}\begin{center}
\includegraphics[width=15cm,height=8cm]{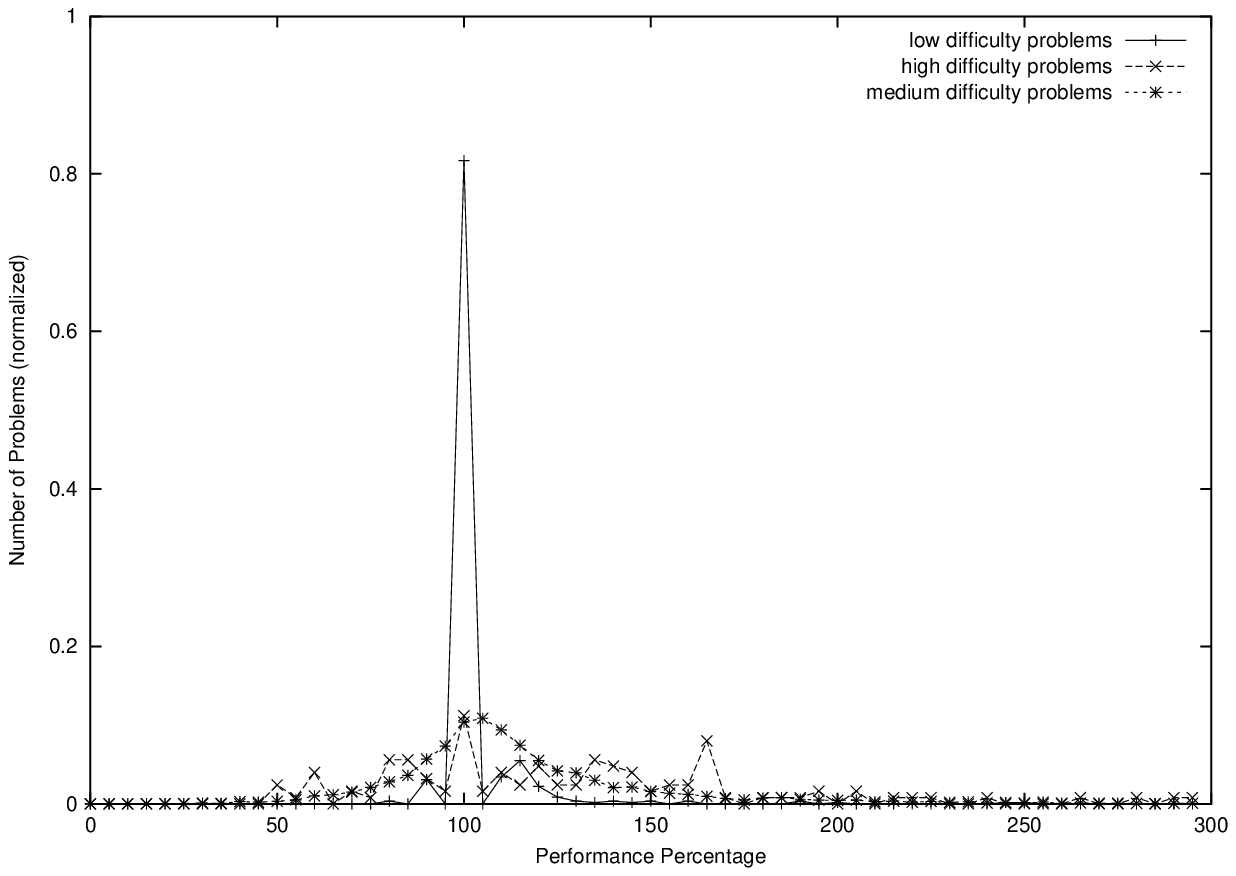}
\includegraphics[width=15cm,height=8cm]{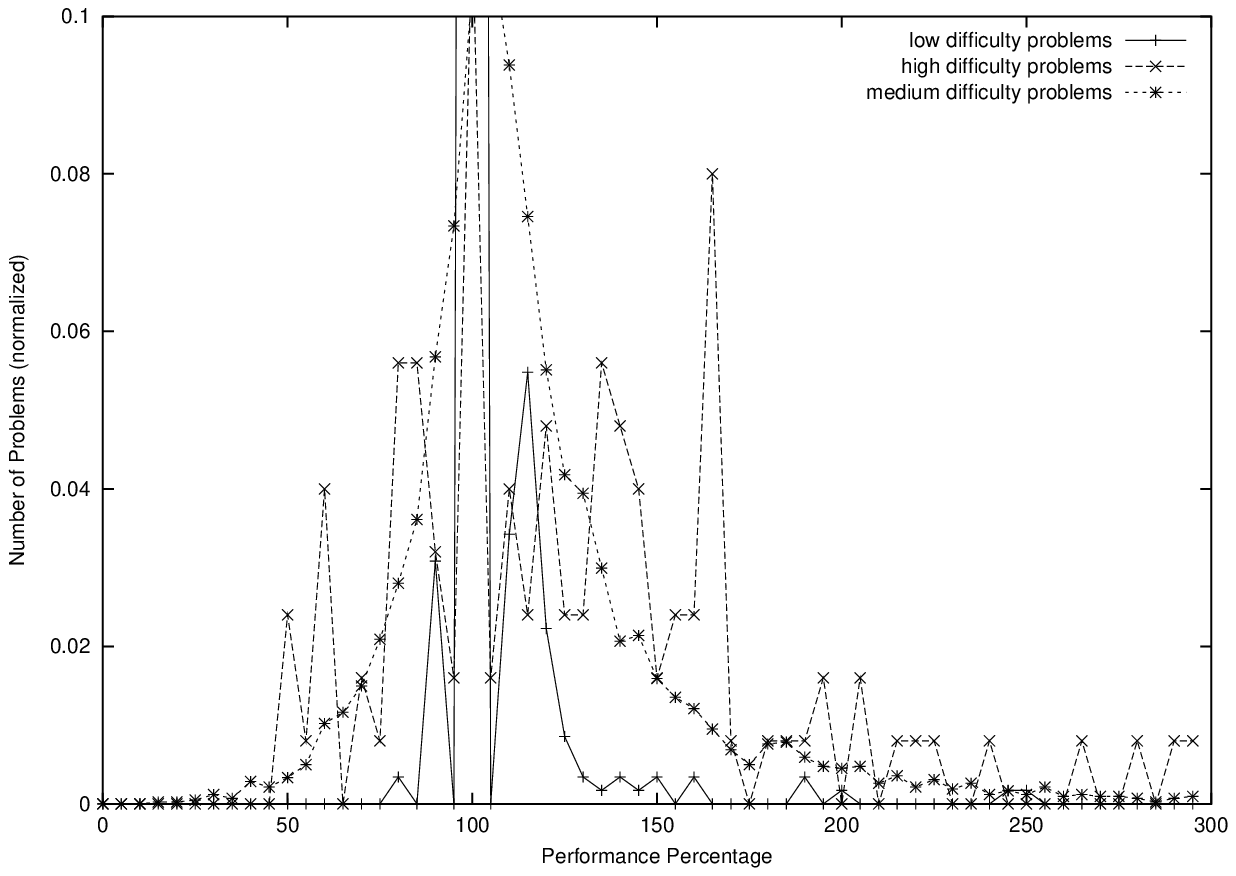}
\caption{Profile plot with static heuristic (trained with static evaluation function) relative to original heuristic for easy, medium and hard difficulty level problems.}
\label{heu-static2-profile}
\end{center}\end{figure}

\begin{figure}\begin{center}
\includegraphics[width=15cm,height=8cm]{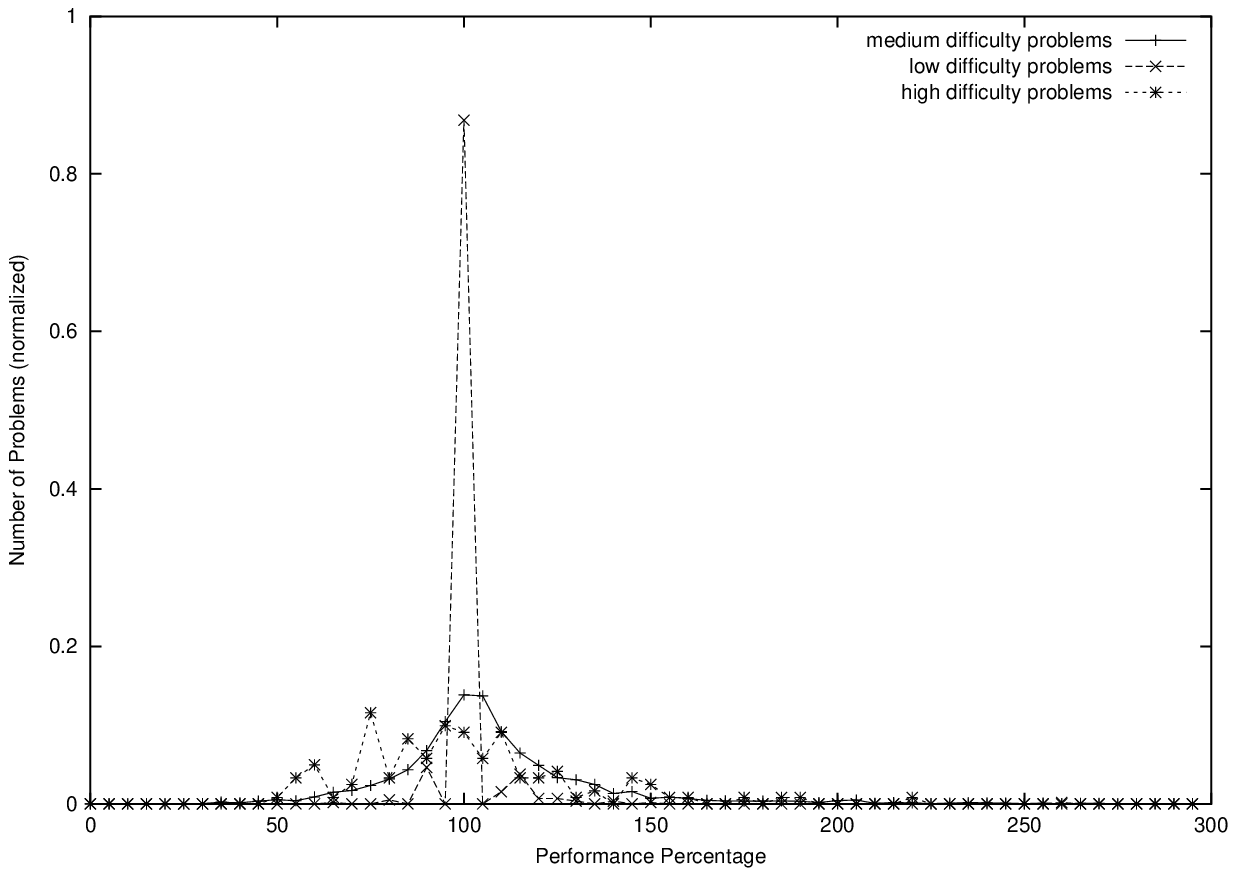}
\includegraphics[width=15cm,height=8cm]{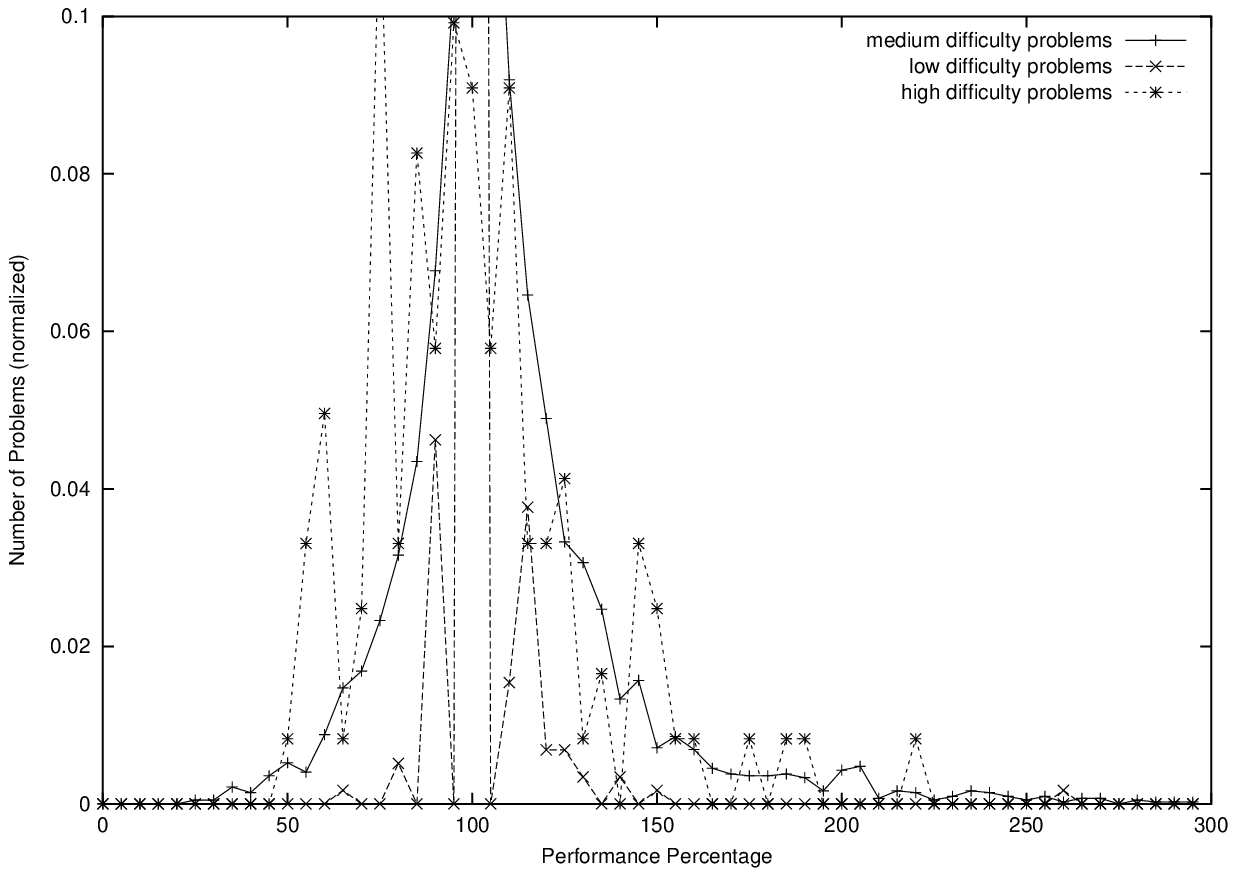}
\caption{Profile plot with static heuristic (trained with dynamic evaluation function) relative to original heuristic for easy, medium and hard difficulty level problems.}
\label{heu-static-solve-profile}
\end{center}\end{figure}

\begin{figure}\begin{center}
\includegraphics[width=15cm,height=8cm]{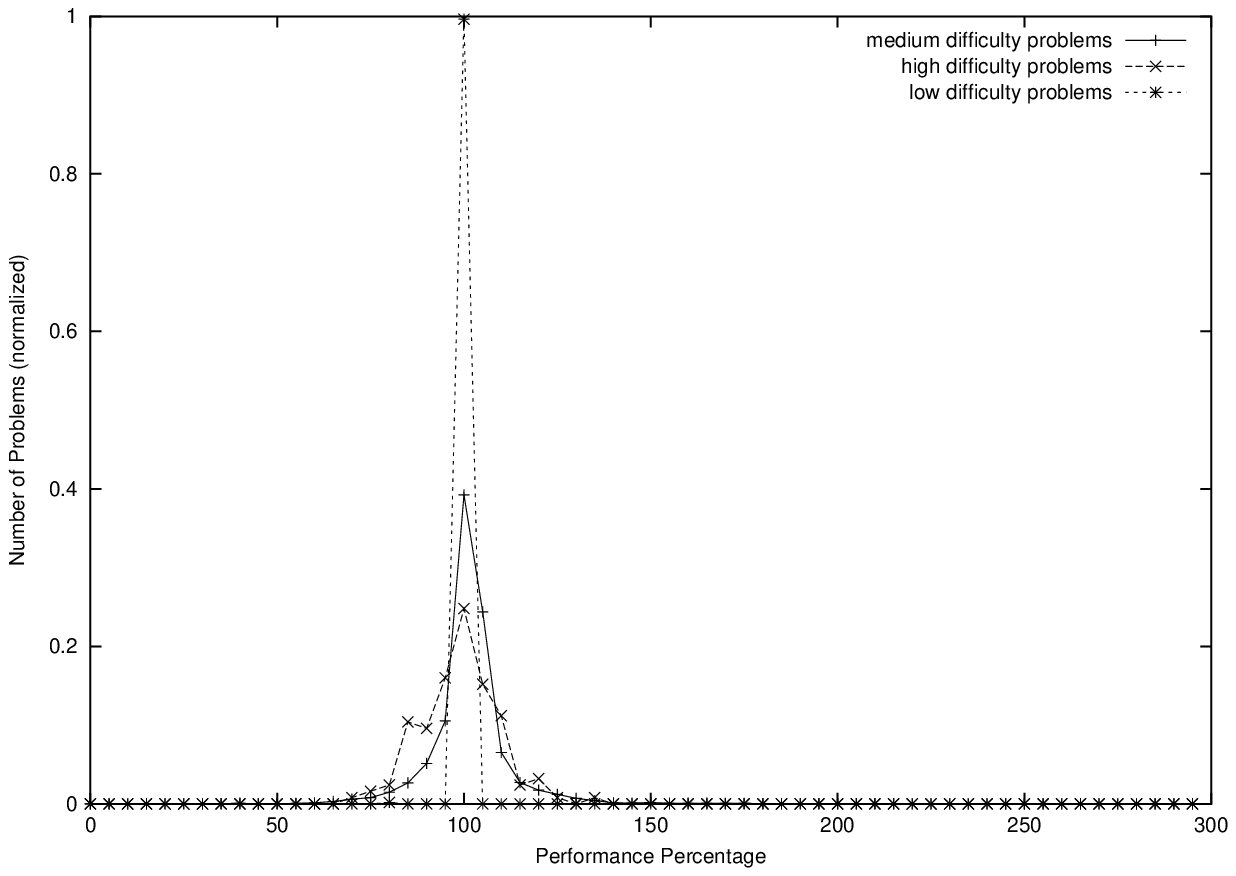}
\includegraphics[width=15cm,height=8cm]{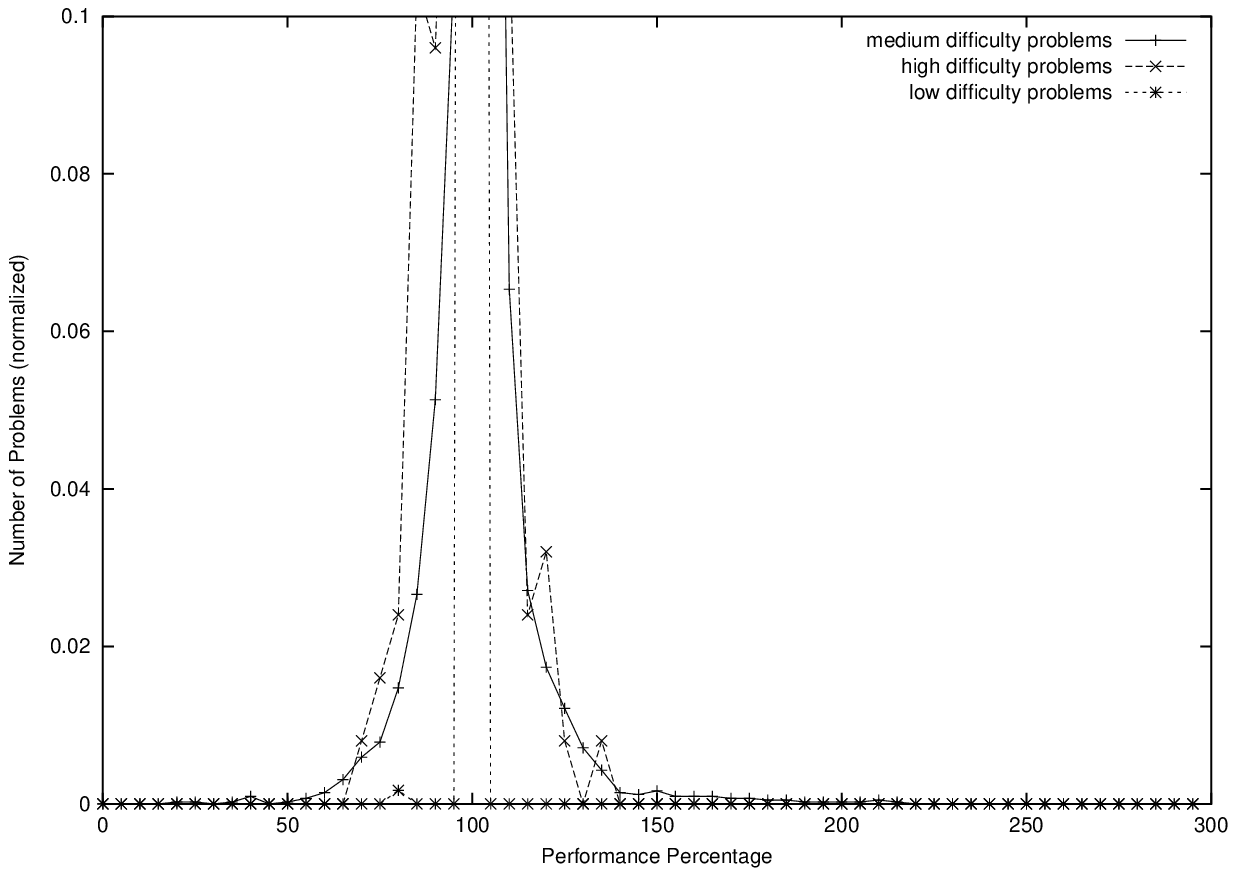}
\caption{Profile plot with dynamic heuristic relative to original heuristic for easy, medium and hard difficulty level problems.}
\label{heu-dynamic-profile}
\end{center}\end{figure}

\begin{figure}\begin{center}
\includegraphics[width=15cm,height=8cm]{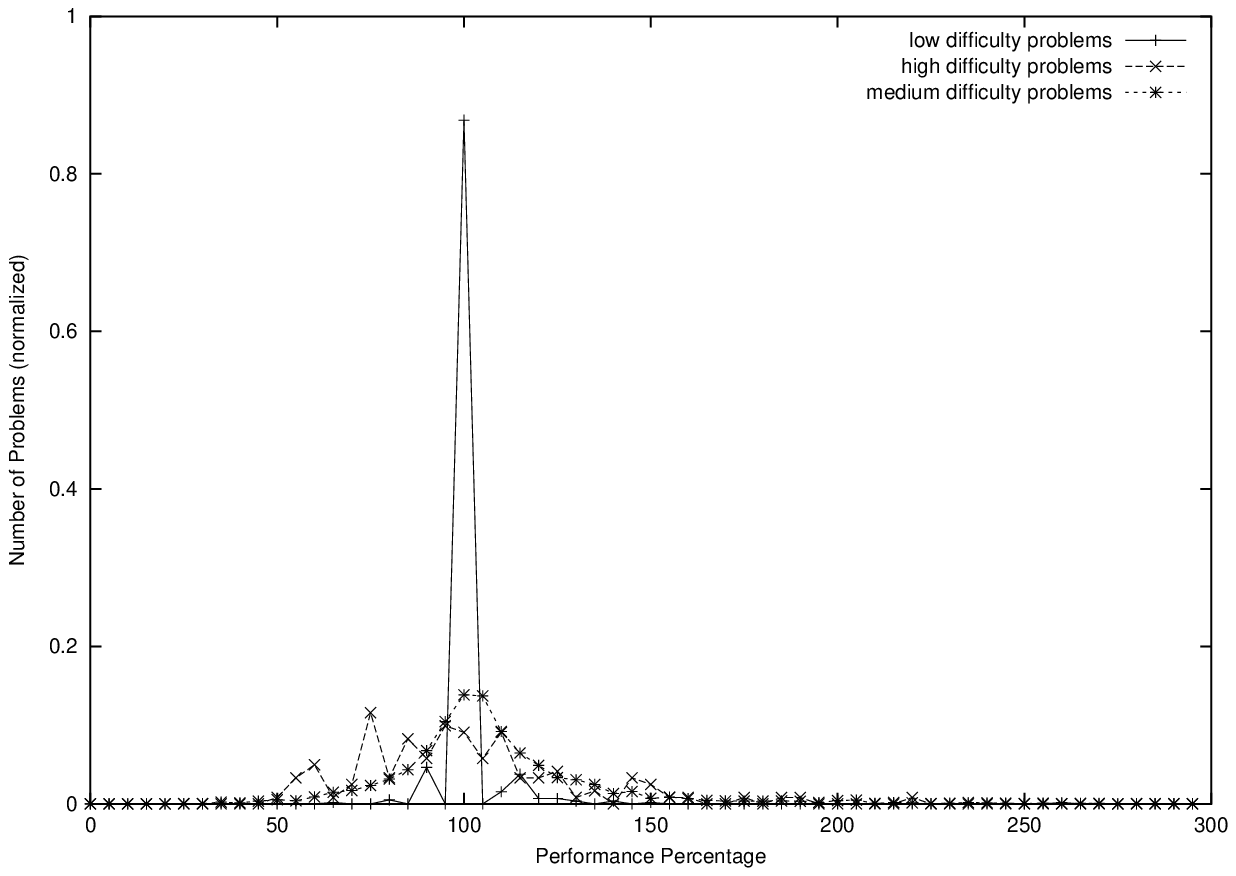}
\includegraphics[width=15cm,height=8cm]{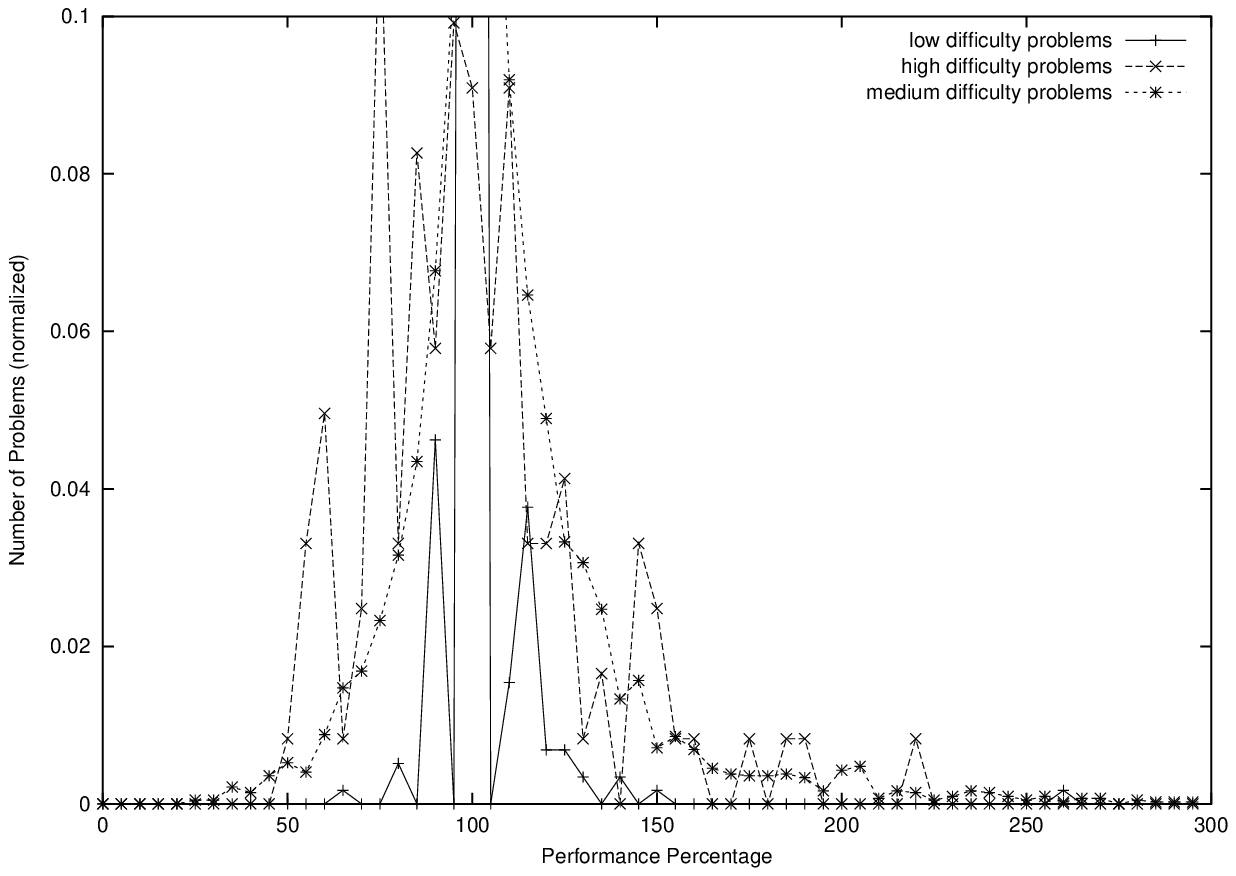}
\caption{Profile plot with best overall (combined) heuristic relative to original heuristic for easy, medium and hard difficulty level problems.}
\label{heu-iamspecial-profile}
\end{center}\end{figure}

To create the profile plots shown in Figures \ref{heu-none-profile} to
\ref{heu-iamspecial-profile}, we ran five heuristics (our baseline,
static weights optimized with the static evaluation function, 
static weights optimized with the dynamic evaluation function, 
dynamic weights and the new optimal (combined) heuristic) relative
to GoTools' existing (i.e.\ previously-optimal) heuristic on our
library of $24,000$ problems.  The x-axis measures the 
new leaf count compared with old leaf count = 100\%.
This means that problems
measuring less than 100\% improved (i.e.\ a reduction in leaf count),
problems measuring greater than 100\% became worse (i.e.\ an increase
in leaf count) while problems close to 100\% showed no performance
change.  The y-axis measures the number of problems counted at a given
performance level, normalized to the range $[0,1]$.  Our graphs are
all plotted in the $x$-range $[0,300]$ once in the y-range $[0,1]$ 
and then again in the y-range $[0,0.1]$ only for better
graphical clarity.  
Finally, we also selected our low, medium and high
problem difficulty levels for the curves shown in each graph.

All of the profile plots shown indicate that very little change occurs
for low difficulty problems.  That is, there is no further
optimization possible for solving low difficulty problems faster given
the current heuristic rules implemented in GoTools. 
The single spike in the plots also indicates that the heuristic rules
are adequate for these problems. As the difficulty
level is increased, the behavior of these five heuristics changes
noticeably.  In Figure \ref{heu-none-profile}, we can see how much
worse GoTools behaves without its heuristic rules as almost all of the
profiled problems took longer to solve.  By contrast, viewing Figure
\ref{heu-static2-profile} confirms that using the static evaluation
function to optimize the static weights resulted in an unstable
profile with some problems being solved faster while most took longer
to solve.  Figure \ref{heu-static-solve-profile} shows improved
results when the static heuristic was trained with the dynamic
evaluation function.  In this case, there is a higher concentration
around the 100\% mark which indicates this heuristic had a higher
consistency across problems than the static heuristic trained with the
static evaluation function.  However, the results are still somewhat
noisy, which is an indication of the limitation of the static
heuristic rules.

The optimized dynamic heuristic shown in Figure
\ref{heu-dynamic-profile} demonstrates the best behavior.  There is
some performance gained as the area below the curves is greater below
the 100\% level than the area above the 100\% level.  Furthermore, the
results indicate a more consistent improvement across many problems
when compared to the variable results seen with the static heuristics.
The overall optimal heuristic shown in Figure
\ref{heu-iamspecial-profile} demonstrates a somewhat combined
behavior which one may expect given that it is formed from the best
dynamic and static heuristic weights learned.  While it does have a
lower consistency than the dynamic heuristics' profile, the overall
improvement still manages to give the best performance of all these
evaluated heuristics.

Looking at the curves for higher difficulty problems the question
arises why the optimized heuristics solve them so inconsistently. One
reason is surely that solving life \& death problems is intrinsically
hard and there is no way to have simple heuristics which are good
enough to solve hard problems without tree-search. On the other hand
it must be said that the heuristic module of GoTools is one of the
weak points of the program. The heuristic rules which currently are mainly
based on superficial issues must be improved to show more
understanding of what the situation is. Only then predictions 
for good moves will be valid for longer sequences of moves and
therefore have value for more difficult problems.

To develop a better heuristic the profiling can be of good use.  By
setting all but one heuristic weight to zero and profiling the
comparison with a run where all weights are zero allows one to filter out
positions where a rule is counterproductive. By filtering out problems
which are solved much slower when solved with one set of heuristic weights versus
another set of weights, one obtains good examples where either individual rules
or combinations of them become counterproductive.

A lesson learned so far is that dynamic rules are more generally
applicable than static rules which seem to loose their predictive
power for difficult problems, at least the rules implemented in
GoTools currently.

\section{Conclusion}

The work described in this paper was meant to improve the current
version of GoTools, as well as laying the groundwork for future
development as the functionality of GoTools is widened to support the
solving of open problems.  The developed tools are:
\begin{itemize} 
\item an MPI interface for Pascal (more specifically an MPICH 
implementation of it), run by us under FreePascal ({\tt
http://www.freepascal.org}) allowing the use of a Beowulf cluster for
genetic learning or any other parallel computation (GoTools +
infrastructure are written in Pascal.),
\item a GA program that can use a static fitness function which computes heuristics, as well
as dynamic fitness functions that solve problem training sets together with large sets of
life \& death problems of varying difficulty.
\end{itemize}

In terms of using these tools to improve the current version
of GoTools, we worked on optimizing 2 sets of weights of heuristic rules.
\begin{description} 
\item[Weights for static rules:] Because these rules take as input only
the current board position we were able to use two different fitness
functions.  One fitness function, called {\em static fitness},
computes a heuristic ranking of all possible moves in a life \& death
problem and gives a bonus according to the place of the unique best
move in this ranking.  The other fitness function, called {\em dynamic
fitness}, solves life \& death problems and takes as fitness measure
the negative of the number of terminal leaves of the search tree.
Both fitness functions operate on a whole training set of problems for
each single chromosome (i.e.\ set of heuristic weights).
Our findings:
\begin{description}
\item Using the static fitness function did not result in a good
training of the static heuristic weights.
\item With the static fitness function, performance on the test set
deteriorated as the problem difficulty was increased.
\item Using the dynamic fitness function, we found that training with
an easy problem set was sufficient to obtain good performance on both
easy and difficult test problems.
\item An improvement of 12\% was realized when using the dynamic
fitness function for our trained static heuristic weights as compared
to zero weight values.
\end{description}
\item[Weights for dynamic rules:]
During the execution of GoTools different forms of learning take
place. One is the filling of a hash data base with the status of
intermediate positions, one is the rule that when backtracking then to
try the other sides winning move first. These forms of learning are
always performed. In addition there are other forms of learning
(how often was a move a winning move in an intermediate position,
negative bonus if a move is forbidden for the enemy,...). The value of
these extra learning heuristics is not so clear cut and therefore
heuristic weights are introduced for them. To optimize these weights
the fitness function has to {\em solve} life \& death problems.
We found that 
\begin{description}
\item A high difficulty training set was sufficient to achieve good performance on easy test sets, and necessary
to achieve good performance on difficult test sets.
\item An improvement of 18\% was realized with our trained dynamic heuristic weights as compared to zero weight values, and
an 8\% improvement over the original dynamic parameter set was achieved with higher difficulty problems.
\item An overall improvement of 20\% was realized when our trained static heuristic weights are combined with our trained dynamic heuristic weights.
\end{description}

\end{description}

\section{Future Work}

The emphasis of the computations done so far was to gather experience
in using Genetic Algorithms to optimize heuristic weights in tsume go.
One can push our results further by using more resources: larger
population sizes, running the GA for more generations and using larger
training sets.  This will definitely be done once the heuristic module
has been overhauled.

In our effort to genetically improve pruning parameters we still have
to gather experience, especially in generating chromosomes evenly
spread over a larger interval of speed-up levels.  However, pruning
parameters depend strongly on the quality of the static and dynamic
heuristics and will become more important in the future when a new
static heuristic module will be completed and be more effective for
difficult open problems.

A more immediate task that emerged from our results is to critically
analyze the different variance values of the optimized individual
heuristic weights, and to check what can be learned about improving
the heuristic rules themselves.

%\bibliography{ga-report-arxiv.bib}

\end{document}